\documentclass[11pt]{article}
\usepackage{bm}
\usepackage{color}
\usepackage{url}
\usepackage{multirow, dsfont}
\usepackage{array}
\usepackage{amsmath}
\usepackage{rotating}
\usepackage{graphicx}
\usepackage{changepage}
\usepackage{sidecap}
\usepackage{makecell}
\usepackage{tikz}
\usetikzlibrary{shapes.geometric, arrows}
\newcolumntype{d}[1]{>{\centering\arraybackslash}m{#1}}
\tikzstyle{item_rec} = [rectangle, rounded corners, minimum width=3cm, minimum height=1cm, text centered, draw=black]
\tikzstyle{item_label} = [rectangle, rounded corners, minimum width=3cm, minimum height=1cm, text centered, draw=black]
\tikzstyle{item_rec1} = [rectangle, rounded corners, minimum width=3.5cm, minimum height=1cm, text centered, draw=black]
\tikzstyle{item_elp} = [ellipse, minimum width=3cm, minimum height=1cm, text centered, draw=black]
\tikzstyle{arrow_t} = [thick, ->, >=stealth]
\tikzstyle{arrow_d} = [dashed, ->, >=stealth]
\tikzstyle{line} = [-stealth, thick, draw]

\DeclareMathOperator*{\argmin}{arg\,\,min\;}
\DeclareMathOperator*{\argmax}{arg\,\,max\;}
\usepackage{caption}

\title{\bf A Novel Minimum Divergence Approach to Robust Speaker Identification}           

\author{Ayanendranath Basu \and Smarajit Bose \and Amita Pal   \and Anish Mukherjee \and Debasmita Das}
\date{}
\begin{document}
\setcounter{table}{0}
\maketitle
\vspace{-1cm}

\begin{center}
\small{Interdisciplinary Statistical Research Unit\\Indian Statistical Institute\\ 
203 B. T. Road, Kolkata 700108, India}

\footnotesize{e-mail: \textit{ayanbasu@isical.ac.in, smarajit@isical.ac.in, pamita@isical.ac.in \\anishmk9@gmail.com, debasmita88@yahoo.com}\\  
}
\end{center}

\begin{abstract}
{\small  In this work, a novel solution to the speaker identification problem is proposed through minimization of statistical divergences between the   probability distribution ($g$) of feature vectors from the test utterance and the probability distributions of the feature vector corresponding to the speaker classes. This approach is made more robust to the presence of outliers, through the use of suitably modified versions of the standard divergence measures. The relevant solutions to the minimum distance methods are referred to as the minimum rescaled modified distance estimators (MRMDEs).  Three measures were considered -- the likelihood disparity, the Hellinger distance and Pearson's  chi-square distance. The proposed approach is motivated by the observation that, in the case of the likelihood disparity, when the empirical distribution function is used to estimate $g$, it becomes equivalent to maximum likelihood classification with Gaussian Mixture Models (GMMs) for speaker classes, a highly effective approach used, for example,  by Reynolds~\cite{reynolds1995} based on Mel Frequency Cepstral Coefficients (MFCCs) as features. Significant improvement in classification accuracy is observed under this approach on the benchmark speech corpus NTIMIT and a new bilingual speech corpus  NISIS, with MFCC features, both in isolation and in combination with delta MFCC features. Moreover, the ubiquitous principal component transformation, by itself and in conjunction with the principle of classifier combination, is found to further enhance the performance.}
\end{abstract}

\thispagestyle{empty}


\section{Introduction}

Automatic speaker identification/recognition (ASI/ASR), that is,  the automated process of inferring the identity of a person from an utterance made by him/her, on the basis of speaker-specific information embedded in the corresponding speech signal, has important practical applications. For example, it can be used to verify identity claims made by users seeking access to secure systems. It has great potential in application areas like voice dialing, secure banking over a telephone network, telephone shopping, database access services, information and reservation services, voice mail, security control for confidential information, and remote access to computers. Another important application of speaker recognition technology is in forensics.

Speaker recognition, being essentially a pattern recognition problem, can be specified broadly in terms of the features used and the classification technique adopted. From experience gained over the past several years from research going on, it has been possible to identify certain groups of features that can be extracted from the complex speech signal, which carry a great deal of speaker-specific information. In conjunction with these features, researchers have also identified classifiers which perform admirably. Mel Frequency Cepstral Coefficients (MFCCs) and Linear Prediction Cepstral Coefficients (LPCCs) are the popularly used features, while Gaussian Mixture Models (GMMs), Hidden Markov Models (HMMs), Vector Quantization (VQ) and Neural Networks are some of the more successful speaker models/classification tools. Any good review article on speaker recognition (for example,~\cite{campbell1997,furui1997,kinnunen2010}), contains details and citations about more than a few of these features and models. It is quite apparent that much of the research involves juggling various features and speaker models in different combinations to get new ASR methodologies. 
Reynolds~\cite{reynolds1995,reynolds1995} proposed a speaker recognition system based on MFCCs as features and GMMs as speaker models and, by implementing it on the benchmark data sets TIMIT~\cite{fisher1986,garofolo1993} and NTIMIT~\cite{garofolo1993}, demonstrated that it works almost flawlessly on clean speech (TIMIT) and quite well on noisy telephone speech (NTIMIT). This successful application of GMMs for modeling speaker identity is motivated by the interpretation that the Gaussian components represent some general speaker-dependent spectral shapes, and also by the capability of mixtures to model arbitrary densities. This approach is one of the most effective approaches available in the literature, as far as accuracy on large speaker databases is concerned. 

In this paper,  a novel approach has been proposed for solving the speaker identification problem through the minimization, over all $K$ speaker classes, of  statistical divergences~\cite{basu2011}  between the (hypothetical) probabilty distribution ($g$) of feature vectors from the test utterance and the probability distribution $f_k$ of the feature vector corresponding to the $k$-th speaker class, $k=1,2,\ldots,K$. The motivation for this approach is provided by the observation that, for one such measure, namely, the Likelihood Disparity, it (the proposed approach) becomes equivalent to the highly successful maximum likelihood classification rule based on Gaussiam Mixture Models for speaker classes~\cite{reynolds1995} with Mel Frequency Cepstral Coefficients (MFCCs) as features. This approach has been made more robust to the possible presence of outlying observations through the use of robustified versions of associated estimators. Three different divergence measures have been considered in this work, and it has been established empirically, with the help of a couple of speech corpora, that the proposed method outperforms the baseline method of Reynolds, when Mel Frequency Cepstral Coefficients (MFCCs) are used as features, both in isolation and in combination with delta MFCC features (Section~\ref{sec:feature}). Moreover, its performance is found to be  enhanced significantly in conjunction with  the following two-pronged approach, which had been shown earlier~\cite{pal2014} to improve the classification accuracy of the basic MFCC-GMM speaker recognition system of Reynolds:

\begin{itemize}
	\item \textit{Incorporation of the individual correlation structures of the feature sets into the model   for each speaker}:  This is a significant aspect of the speaker models that Reynolds had ignored by assuming the MFCCs to be independent. In fact, this has given rise to the misconception that MFCCs are uncorrelated. Our objective is achieved by the simple device of the Principal Component Transformation (PCT)~\cite{rao2001}. This is a linear transformation derived from the covariance matrix of the feature vectors obtained from the training  utterances of a given speaker, and is  applied to the feature vectors  of the corresponding speaker to make the individual coefficients uncorrelated. Due to differences in the correlation structures, these transformations are also different for different speakers. The GMMs are fitted on the feature vectors transformed by the principal component transformations rather than the original featuress. For testing, to determine the likelihood values with respect to a given target speaker model, the feature vectors computed from the test utterance are rotated by the principal component transformation corresponding to that speaker.
	\item \textit{Combination of  different classifiers based on the MFCC-GMM model: }
Different classifiers are built by varying some of the parameters of the model. The performance of these classifiers in terms of classification accuracy also varies to some extent. By combining the decisions of these classifiers in a suitable way, an aggregate classifier is built whose performance is better than any of the constituent classifiers. 
\end{itemize}

The application of Principal Component Analysis (PCA) is certainly not new in the domain of speaker recognition, though the primary aim has been to implement dimensionality reduction~\cite{chien2004,hanilci2009,seo2001,suri2013,vijendra2013,zhang2003} for improving performance.  The novelty of the approach used here (proposed by Pal \textsl{et al}.~\cite{pal2014} lies in the fact that   the principle underlying PCA  has been used to make the features uncorrelated, without trying to reduce the size of the data set. To emphasize this feature, we refer to our implementation as the Principal Component Transformation (PCT) and not PCA. Moreover, another unique feature of our approach is as follows. We compute the PCT for each speaker on the training utterances and store them. GMMs for a speaker are estimated based on the feature vectors transformed by its PCT. For testing, unlike what has been reported in other work, in order to determine the likelihood values with respect to a given target speaker model, the MFCCs computed from the test utterance are rotated by the PCT for that target speaker, and not the PCT determined from the test signal itself. The motivation is that if the test signal comes from this target speaker, when transformed by the corresponding PCT, it will match the model better.

The principle of combination or aggregation of classifiers for improvement in accuracy has been used successfully in the past for speaker recognition, for example, by Besacier and Bonastre~\cite{besacier2000}, Altin\c{c}ay and Demirekler~\cite{altincay2003}, Hanil\c{c}i and Erta\c{s}~\cite{hanilci2009}, Trabelsi and Ben Ayed~\cite{trabelsi2013}. In the approach proposed in this work, different type of classifiers are not combined. Rather,  a few GMM-based classifiers are  generated  and their decisions are combined. This is somewhat similar to the principle of \textit{Bagging}~\cite{breiman1996} or \textit{Random Forests}~\cite{breiman2001}.

The proposed approach has been implemented on the benchmark speech corpus, NTIMIT, as well as a relatively new bilingual speech corpus NISIS~\cite{pal2012}, and  noticeable improvement in recognition performance is observed in both cases, when Mel Frequency Cepstral Coefficients (MFCCs) are used as features, both in isolation and in combination with delta MFCC features. 

The paper is organized as follows. The minimum distance (or divergence) approach is introduced in the following section, together with a few divergence measures. The proposed approach is presented in  Section 3, which also outlines the motivation for it.  Section 4 gives a brief description of the speech corpora used, namely, NISIS and NTIMIT, and contains results obtained by applying the proposed approach on them, which clearly establish its effectiveness. Section 5 summarizes the contribution of this work and proposes future directions for research in this area.

\section{Divergence Measures}

Let $f$ and $g$ be two probability density functions. Let the Pearson's residual~\cite{lindsay1994} for $g$, relative to $f$, at the value $\bm{x}$ be defined as $$ \delta(\bm{x}) = \frac{g(\bm{x})}{f(\bm{x})} - 1. $$ 

The residual is equal to zero at such values where the densities $g$ and $f$ are identical. We will consider divergences between $g$  and $f$ defined by the general form  
\begin{equation}
\label{eq:general}
\rho_C(g,f) = \int_{\bm{x}} C(\delta(\bm{x}))\,f(\bm{x})\,{d\bm{x}}, 
\end{equation}
 where $C$ is a thrice differentiable, strictly convex function on $[-1, \infty)$, satisfying $C(0) = 0$. 

Specific forms of the function $C$ generate different divergence measures. In particular, the likelihood disparity (LD) is generated when $C(\delta) \; = \; (\delta+1) \, \log(\delta+1) \; - \; \delta$. Thus, $$ LD(g,f) = \int_{\bm{x}} [(\delta(\bm{x})+1) \, \log(\delta(\bm{x})+1) \; - \; \delta(\bm{x})] \; f(\bm{x}) \, d\bm{x} $$ which ultimately reduces upon simplification to
\begin{equation}
 LD(g,f) = \int_{\bm{x}} \log(\delta(\bm{x})+1) \, dG = \int_{\bm{x}} \log(g(\bm{x}))\, dG \; - \int_{\bm{x}} \log(f(\bm{x}))\,dG,
\end{equation}
where $G$ is the distribution function corresponding to $g$.
For the Hellinger distance (HD), since $ C(\delta) = 2(\sqrt{\delta+1}-1)^2 $, we have $$ HD(g,f) = 2\int_{\bm{x}} \big(\sqrt{\frac{g(\bm{x})}{f(\bm{x})}} - 1\big)^2 f(\bm{x}) \, d\bm{x}, $$ which can be expressed (upto an additive constant independent of $g$ and $f$) as 
\begin{equation}
\label{eq:3}
HD(g,f) = -4\int_{\bm{x}} \frac{1}{\sqrt{\delta(\bm{x}) + 1}} \, dG.
\end{equation} 
For Pearson's chi-square (PCS) divergence, $ C(\delta) = \delta^2/2 $, so $$ PCS(g,f) = \frac{1}{2}\int_{\bm{x}} \big(\frac{g(\bm{x})}{f(\bm{x})} - 1\big)^2 f(\bm{x}) \, d\bm{x}, $$ which simplifies (upto an additive  constant independent of $g$ and $f$) to
\begin{equation} 
\label{eq:4}
PCS(g, f) = \frac{1}{2}\int_{\bm{x}} \big(\delta(\bm{x})+1\big) \, dG.
\end{equation}
The divergences within the general class described in (\ref{eq:general}) have been called disparities~\cite{basu2011,lindsay1994}. The LD, HD and the PCS denote three prominent members of this class.

\subsection{Minimum Distance Estimation}
Let $X_1, X_2,\ldots,X_n$ represent a random sample from a distribution $G$ having a probability density function
$g$ with respect to the Lebesgue measure.  Let $\hat{g}_n$
represent a density estimator of $g$  based on the random sample.  Let the parametric model family $\cal F$, which models
the true data-generating distribution $G$, be defined as ${\cal F}=\{F_\theta:\theta \in \Theta \subseteq I\!\!R^p\}$, where $\Theta$ is the parameter space. Let  $\cal G$ denote the class of all distributions having densities with respect to the
Lebesgue measure,  this class being assumed to be convex. It is further assumed  that both the data-generating distribution $G$ and the model family
$\cal F$ belong to $\cal G$. Let $g$ and $f_{\theta}$ denote the probability density functions corresponding to $G$ and $F_{\theta}$. Note that $\theta$ may represent a continuous parameter as in usual parametric inference problems of statistics, or it may be discrete-valued, if it denotes the class label in a classification problem like speaker recognition.

The minimum distance estimation approach for estimating the parameter $\theta$  involves the determination the element of the model family
which provides the closest match to the data in terms of the distance (more generally, divergence) under
consideration.
That is, the minimum distance estimator $\hat{\theta}$ of $\theta$ based on the divergence $\rho_C$ is defined by the relation 
$$\rho_C(\hat{g}_n,f_{\hat{\theta}})=\min_{\theta \in \Theta} \rho_C(\hat{g}_n,f_{\theta}).$$

When we use the likelihood disparity (LD) to assess
the closeness between the data and the model densities, we determine the element $f_\theta$ which is closest to $g$ in terms of the likelihood disparity. In this case the procedure, as we have seen in Equation (\ref{eq:LD}), becomes equivalent to the choice of the element $f_\theta$ which maximizes 
$\int_{\bm{x}} \log(f_\theta(\bm{x}))\,dG(\bm{x}).$ As $g$ (and the corresponding distribution function $G$) is unknown, we need to optimize a sample based version of the objective function. While in general this will require the construction of a kernel density estimator $\hat{g}$ (or an alternative density estimator), in case of the likelihood disparity this is provided by simply replacing the differential $dG$ with 
$dG_n$, where $G_n$ is the empirical distribution function. The procedure based on the minimization of the objective function in Equation (2) then further simplifies to the maximization of 
$$\frac{1}{n} \sum_{i=1}^n \log f_\theta(X_i)$$
which is equivalent to the maximization of the log likelihood.

The above demonstrates a simple fact, well-known in the density-based minimum distance literature or in information theory, but not well-perceived by most scientists including many statisticians: the maximization of the log-likelihood is equivalently a minimum distance procedure. This provides our basic motivation in this paper. Although we base our numerical work on the three divergences considered in the previous section, our primary intent is to study the general class of minimum distance procedures in the speech-recognition context such that the maximum likelihood procedure is a special case of our approach. Many of the other divergences within the class generated by 
Equation (\ref{eq:general}) also  have equivalent objective functions that are to be maximized to obtain the solution and have simple interpretations. 

However, in one respect the likelihood disparity is unique. It is the only divergence in this class where the sample based version of the objective function may be created by the simple use of the empirical and no other nonparametric density estimation is required. Observe that both in Equations (\ref{eq:3}) and (\ref{eq:4}), the integrand involves $\delta(\bm{x})$, and therefore a density estimate for $g$ is required even after replacing $dG$ by $dG_n$.  
 
\subsection{Robustified Minimum Distance Estimators}
When the divergence $\rho_C(\hat{g}_n, f_\theta)$ is  differentiable with respect to $\theta$, the minimum distance estimator $\hat{\theta}$ of $\theta$ based on the
divergence $\rho_C$ is obtained by solving the estimating equation
\begin{equation}
-\nabla \rho_C(\hat{g}_n,f_{\theta})=\int_x A(\delta(x))\nabla f_\theta (x) dx=0,
\end{equation}
where the function $A(\delta)$ is defined as $$A(\delta)=C'(\delta)(\delta+1)-C(\delta).$$ 
If the function $A(\delta)$  satisfies
$A(0) = 0$ and $A'(0) = 1$ then it is termed  the Residual Adjustment Function (RAF) of the divergence. Here $\nabla$ denotes the gradient operator with respect to $\theta$, and $C'(\cdot)$ and $A'(\cdot)$ represent the respective derivatives of the functions $C$ and $A$ with respect to their arguments.

Since the estimating equations of the different minimum distance estimators differ only in the form of the residual adjustment
function $A(\delta)$, it follows that the  properties of these  
estimators must be determined by the form of the corresponding function $A(\delta)$. Since
$A'(\delta) = (\delta + 1)C''(\delta)$ and, as $C(\cdot)$ is a strictly convex function on $[-1, \infty)$, $A'(\delta) > 0$ for
$\delta > −1$; hence $A(\cdot)$ is a strictly increasing function on $[−1,\infty)$.

Geometrically, the RAF is the most important tool to demonstrate the general behaviour or the heuristic robustness properties of the minimum distance estimators corresponding to the class defined in (\ref{eq:general}).
A dampened response to increasing positive $\delta$ will ensure that the RAF shrinks
the effect of large outliers as $\delta$ increases, thus providing a strategy for making the corresponding minimum distance estimator robust to outliers.

For the likelihood disparity (LD), $C(\delta)$ is unbounded for large positive values of the residual $\delta$. and the corresponding estimating equation is given by, $$ -\nabla\: LD(g,f_\theta) = \int_{\bm{x}} \delta\nabla f_\theta = 0. $$ So, the residual adjustment function (RAF) for LD, $A_{LD}(\delta) = \delta$, increases linearly in $\delta$. Thus, to dampen the effect of outliers,   a modified $A(\delta)$ function could be used, which is defined as
\begin{equation}
A(\delta) = 
     \begin{cases}
       0 &\quad\text{for } \; \delta \in [-1, \alpha] \cup [\alpha^*, \infty); \\
       \delta &\quad\text{for } \; \delta \in (\alpha, \alpha^*).
     \end{cases}
\end{equation}
This eliminates the effect of large $\delta$ residuals  beyond the range $(\alpha, \alpha^*)$. This proposal is in the spirit of the trimmed mean.

The $C(\delta)$ function for the modified LD (MLD) reduces to
\begin{equation}
C_{MLD}(\delta) = 
     \begin{cases}
       0 &\quad\text{for } \; \delta \in [-1, \alpha] \cup [\alpha^*, \infty); \\
       (\delta+1)\log(\delta+1) - \delta &\quad\text{for } \; \delta \in (\alpha, \alpha^*).
     \end{cases}
\end{equation}

Similarly, the RAF for the Hellinger distance is $ A_{HD} = 2(\sqrt{\delta+1} - 1) $, which too is unbounded  for large values of $\delta$, in spite of its local robustness properties. To obtain a robustified estimator,  the RAF is modified to
\begin{equation}
A(\delta) = 
     \begin{cases}
       0 &\quad\text{for } \; \delta \in [-1, \alpha] \cup [\alpha^*, \infty); \\
       2(\sqrt{\delta+1} - 1) &\quad\text{for } \; \delta \in (\alpha, \alpha^*),
     \end{cases}
\end{equation}
so that the $C(\delta)$ function for the modified HD (MHD) becomes
\begin{equation}
C_{MHD}(\delta) = 
     \begin{cases}
       0 &\quad\text{for } \; \delta \in [-1, \alpha] \cup [\alpha^*, \infty); \\
       2(\sqrt{\delta+1}-1)^2 &\quad\text{for } \; \delta \in (\alpha, \alpha^*).
     \end{cases}
\end{equation}

For Pearson's chi-square (PCS) divergence, $A(\delta) = \delta + \frac{\delta^2}{2}$ is again unbounded for large $\delta$, so the RAF is modified to
\begin{equation}
A(\delta) = 
     \begin{cases}
       0 &\quad\text{for } \; \delta \in [-1, \alpha] \cup [\alpha^*, \infty); \\
       \delta + \frac{\delta^2}{2} &\quad\text{for } \; \delta \in (\alpha, \alpha^*),
     \end{cases}
\end{equation}
so that the $C(\delta)$ function for the modified PCS (MPCS) becomes
\begin{equation}
C_{MPCS}(\delta) = 
     \begin{cases}
       0 &\quad\text{for } \; \delta \in [-1, \alpha] \cup [\alpha^*, \infty); \\
       \frac{\delta^2}{2} &\quad\text{for } \; \delta \in (\alpha, \alpha^*).
     \end{cases}
\end{equation}
In Figure 1, we have presented the RAFs of our three candidate divergences, the LD, the HD and the PCS. Notice that they have three different forms. The RAF of the LD is linear, that of the HD is concave, while the PCS has a convex RAF. We have chosen our three candidates as representatives of these three types, so that we have a wide description of the divergences of the different types.

\begin{figure}[htb!]
\begin{center}
\includegraphics[scale=0.64]{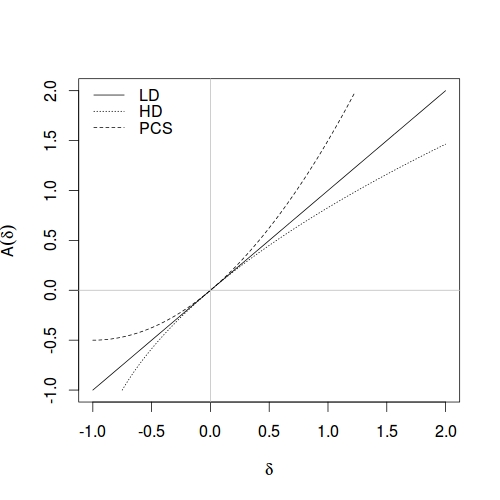}
\caption{The Residual Adjustment Functions (RAFs) of the LD, HD and PCS divergences}
\end{center}

\end{figure}

\textbf{Remark 1}: In the above proposals, the approach to robustness is not through the intrinsic behaviour of the divergences, but through the trimming of highly discordant residuals. For small-to-moderate residuals, the RAFs of these divergences are not widely different, as all of them relate to the treatment of residuals which do not exhibit extreme departures from the model. However, these small deviations often provide substantial differences in the in the behavior of the corresponding estimators. We hope to find out how the small departures exhibited in these divergences are reflected in their classification performance.

\textbf{Remark 2}: In this paper, our minimization of the divergence will be over a discrete set corresponding to the indices of the existing speakers in the database that the new utterance is matched against. Thus we will not directly use the estimating equation in (5) to ascertain the minimizer. In fact if we restrict ourselves just to the three divergences considered here, there would be no reason to use the residual adjustment function. However these divergences are only representatives of a bigger class, and generally the properties of the minimum distance estimators are best understood through residual adjustment function. Reconstructing the function $C(\cdot)$ from the residual adjustment function $A(\cdot)$ requires solving an appropriate differential equation. When this reconstruction does not lead to a closed form of the $C(\cdot)$, one has to directly use the form of the residual adjustment function for the minimizations considered in this paper. 

\textbf{Remark 3}: Any divergence of the form described in Equation (1) can be expressed in terms of several distinct $C(\delta)$ functions. While they lead to the same divergence when integrated over the entire space,  when the range is truncated by eliminating very large and very small residuals, the role of the $C(\cdot)$ function becomes important. In this section we have modified the likelihood disparity, the Hellinger distance and the Pearson's chi-square by truncating the $C(\cdot)$ functions having the form
$$
C_{LD}(\delta)  =  (\delta +1) \log (\delta + 1) - \delta,\;\;
C_{HD}  = 2(\sqrt{\delta +1} - 1)^2,\;\;
C_{PCS}  =  \frac{\delta^2}{2}.
$$
One could also modify the versions presented in Equations (2), (3) and (4) in a similar spirit and obtain truncated solutions of the minimization problem under study. 

\section{The Proposed Approach}
\label{sec:proposed}

It is assumed that probability distribution $g$ for the (unknown) speaker of the test utterance is unknown. However, it can be estimated by $\hat{g}$ computed from the test utterance using the feature vectors $\bm{x}_i$'s, corresponding to a number of overlapping short-duration segments into which the segment can be divided. The proposed approach aims to identify $k^*$, for which $f_{k^*}$ is \textit{most similar} to $g$ in the \textit{minimum distance} sense, where $f_k,\:k=1,2,\ldots,K$, are the probability models for the $K$ speaker classes.  In other words, the proposed approach infers that speaker number $k^*$ has uttered the test speech if $$ k^* = \argmin_k \rho_C(g,f_k), $$ where $\rho_C(\cdot,\cdot)$ is some statistical divergence measure between two probability density functions, for a given choice of the function $C$.
If the Pearson's residual for $g$ relative to $f_k$ at the value $\bm{x}$ be defined by $$ \delta_k(\bm{x}) = \frac{g(\bm{x})}{f_k(\bm{x})} - 1, $$ then the divergence between $g$ and $f_k$ is given by $$ \rho_C(g, f_k) = \int_{\bm{x}} C(\delta_k(\bm{x}))\,f_k(\bm{x})\,{d\bm{x}}. $$  

Let $\bm{X}_1,\bm{X}_2, \ldots, \bm{X}_M$ be a random sample of size $M$ from $g$ and let us estimate the corresponding distribution function $G$ by the empirical distribution function  $$ G_n(\bm{x}) = \frac{1}{M} \sum\limits_{i = 1}^{M} \mathds{1}_{(X_i \, \leq \; \bm{x})} $$  based on the data $\bm{x}_i, \; i = 1,\ldots,M$, where $\mathds{1}_{(A)}$ is the indicator of the set $A$.
\subsection{Modified Minimum Distance Estimation}
As noted earlier, specific forms of the function $C(\cdot)$ generate different divergence measures. In the following, we will describe the identification of the speaker of the test utterance based on the three divergencees considered in Section 2.

\subsubsection{Estimation based on the Likelihood Disparity}
The likelihood disparity (LD) between $g$ and $f_k$ is (upto an additive constant)
\begin{equation}
\label{eq:LD}
 LD(g,f_k) = \int_{\bm{x}} \log(\delta_k(\bm{x})+1) \, dG = \int_{\bm{x}} \log(g(\bm{x}))\, dG \; - \int_{\bm{x}} \log(f_k(\bm{x}))\,dG,
\end{equation}
Under the proposed approach, the speaker of a test utterance is identified by minimizing the likelihood disparity between $g$ and the $f_k$'s, that is, as speaker number $k^*$ if $$ k^* = \argmin_k LD(g,f_k) = \argmax_k \int_{\bm{x}} \log(f_k(\bm{x}))\,dG $$ where the second equality holds because the first term in the expression of $LD(g,f_k)$ given in Equation (12) does not involve $f_k$.  Since $\int_{\bm{x}} \log(f_k(\bm{x}))\,dG_n$ is an estimator of $\int_{\bm{x}} \log(f_k(\bm{x}))\,dG$, we have 

\begin{equation}
\label{eq:loglik}
\int_{\bm{x}} \log(f_k(\bm{x}))\,dG \approx \int_{\bm{x}} \log(f_k(\bm{x}))\,dG_n = \frac{1}{M} \sum\limits_{i = 1}^M \log(f_k(\bm{x_i})).
\end{equation}
Therefore, we will choose the index by maximizing the log-likelihood, which gives 
\begin{equation}
\label{mle}
\hat{k}^* = \argmax_k \sum\limits_{i = 1}^M \log(f_k(\bm{x_i})) = \argmax_k \prod\limits_{i = 1}^M f_k(\bm{x_i}).
\end{equation}

\subsubsection{Estimation based on the Hellinger Distance}
Using the form described in Equation (\ref{eq:3}), the  Hellinger distance (HD)  between $g$ and $f_k$ is the same (upto an additive constant) as
\begin{equation} 
\label{eq:HD}
HD(g, f_k) = -4\int_{\bm{x}} \frac{1}{\sqrt{\delta_k(\bm{x}) + 1}} \, dG.
\end{equation} 
By the same reasoning as before, the speaker of the test utterance is determined to be speaker number $k^*$, by minimizing the empirical version of the Hellinger distance between $g$ and $f_k$'s, that is,
\begin{equation}
\hat{k}^* = \argmax_k \sum\limits_{i = 1}^M \frac{1}{\sqrt{\delta_k(\bm{x_i})+1}}.
\end{equation}
We have dropped the factor of $1/M$ as it has no role in the maximization. However in this case we have to substitute a density estimate of $g$ in the 
expression of $\delta_k$. Here we will do this using a Gaussian mixture model. 
\subsubsection{Estimation based on the Pearson Chi-square Distance}
Using the form described in Equation~(\ref{eq:4}), the Pearson's chi-square between $g$ and $f_k$ is the same as (up to an additive constant) 
\begin{equation} 
\label{eq:PCS}
PCS(g, f_k) = \frac{1}{2}\int_{\bm{x}} \big(\delta_k(\bm{x})+1\big) \, dG.
\end{equation}
Thus, as before, speaker number $k^*$ is identified as having produced the test utterance if
\begin{equation}
\hat{k}^* = \argmin_k \sum\limits_{i = 1}^M \big(\delta_k(\bm{x_i})+1\big).
\end{equation}
For each of the three divergences considered in Sections 3.1.1-3.1.3, we trim the empirical versions of the divergences in the spirit of Section 2.2. This will mean that our modified objective function for the three divergences (LD, HD and PCS)) are, respectively 
$$\sum_{i \in B} \log f_k(\bm{x}_i)
, ~~ \sum_{i \in B} \frac{1}{\sqrt{\delta_k(\bm{x}_i)+1}}, ~{\rm and}~ \sum_{i \in B} (\delta_k(\bm{x}_i)+1),$$
where the set $B$ may be defined as $B = \{i| \delta_k(\bm{x}_i) \in (\alpha, \alpha^*)\}$; the set $B$ depends on $k$ also, but we keep the dependence implicit. In our experimentation, we have varied both $\alpha$ and $\alpha^*$ in order to control the effect of both outliers and inliers and chose the  pair that led to  maximum speaker identification accuracy.
\subsection{Minimum Rescaled Modified  Distance Estimation}
In our implementation of the above proposal, we chose $\alpha$ and $\alpha^*$ not as absolutely fixed values, but as values which will provide a fixed level of trimming (like 10\% or 20\%). However, on account of the very high dimensionality of the data and the availability of a relatively small number of data points for each test utterance, the estimated densities are often very spiky, leading to very high estimated densities at the observed data points. This, in turn, often leads to very high Pearson residuals at such observations. Since the choice of the tuning parameters is related to the trimming of a fixed proportion of observations, many of the untrimmed observations may still be associated  with very high Pearson residuals, which makes the estimation unreliable.    As a result, $\delta$ becomes very large at a  majority of the sample points of the test utterances, which impacts heavily on the divergence measures. 

From~(\ref{eq:LD}) we see that, $\delta_k(\bm{x}_i), \; i = 1,\ldots,M$ are in logarithmic scale in the expression of LD. In fact Equation (13) shows that the final objective function in case of the empirical version of the likelihood disparity does not directly depend on the values of the Pearson residuals at all. Thus, although $\delta_k(\bm{x}_i)$ values are large, LD gives quite sensible divergence values. But, in case of the HD as given in Eq.~(\ref{eq:HD}) and the PCS as given in Eq.~(\ref{eq:PCS}), we find that the divergence values are greatly affected by the large $\delta_k(\bm{x}_i)$ values for majority of $i$'s. Thus, in order to reduce the impact of large $\delta$ values on the HD and PCS, we propose a scaled version of the residual  $\delta$ as follows:
\begin{equation}
\delta^* = \mbox{sign}(\delta) \; |\delta|^\beta
\end{equation}
where 
$$
\mbox{sign}(\delta) = 
     \begin{cases}
       1 &\quad\text{for } \; \delta \geq 0, \\
       -1 &\quad\text{for } \; \delta < \,0.
     \end{cases} 
$$
and $\beta$ is a positive scaling parameter which can be used to control the impact of $\delta$.
For a value of $\beta$ significantly smaller than 1, $\delta^*$ is scaled down to a much smaller value in magnitude compared to $\delta$. With this modification, then, our relevant objective functions for the LD, HD and PCS are 
 $$\sum_{i \in B} \log f_k(\bm{x}_i)
, ~~ \sum_{i \in B} \frac{1}{\sqrt{\delta^*_k(\bm{x}_i)+1}}, ~{\rm and}~ \sum_{i \in B} (\delta^*_k(\bm{x}_i)+1).$$
Notice that the objective function for LD remains the same as described in Section 3.1, but the objective functions for the HD and PCS are the same only when $\beta = 1$. 

We will refer to the estimators obtained by minimizing the rescaled, modified objective functions as the Minimum Rescaled Modified  Distance Estimators (MRMDEs) of type I. Only in case of the likelihood disparity the rescaling part is absent. 

\subsection{Minimum Rescaled Modified  Distance Estimators (MRMDEs) of Type II}
In the previous subsection we have described the construction of the MRMDEs of type I. In Remark 3 we have mentioned that the same divergence may be constructed by several distinct $C(\cdot)$ functions. While they provide identical results when integrated over the entire space, the modified versions corresponding to the different $C(\cdot)$ functions are necessarily different, although the differences are often small. 

Note that 
$$\int C(\delta_k(\bm{x})) f_k(\bm{x}) d\bm{x} 
 =  \int \frac{C(\delta_k(\bm{x})}{(\delta_k(\bm{x})+1}
dG(\bm{x}),$$
and using the same principles as in Sections 3.1 and 3.2, 
we propose the minimization of the objective function
$$\sum_{i \in B} \frac{C(\delta^*(\bm{x}_i))}{(\delta_k^*(\bm{x}_i)+1)}$$
for the evaluation of the MRMDEs of Type II. Here the relevant $C(\cdot)$
functions corresponding to the LD, HD and PCS are as defined in Equations
(7), (9) and (11). Note that in this case the rescaling has to be applied to all 
the three divergences, and not just to HD and PCS only.

\section{The Principal Component Transformation}
The idea of principal component transformation (PCT) as proposed in an earlier work~\cite{pal2014} has also been used here. Let the PCT matrix of $k^{th}$ speaker be $P_k, \; k = 1,\ldots,K$ and $X_k(d \times M_k)$ be the training feature matrix for $k^{th}$ speaker, where $d$ = dimension of feature vector and $M_k$ = number of feature vectors. In the training phase, we first get the transformed feature matrix $X_k^*$ as, 
\begin{equation}
\label{eq:PCT}
X^*_k = P_k X_k
\end{equation} 
and then use it to train $f_k$. Now in the testing phase, we extract the feature matrix from a test utterance represented by $X$, compute the PCT matrix $P$ and obtain the transformed feature matrix $X^*$ as in~(\ref{eq:PCT}). Then we train the model $g$ using $X^*$.

Let us define $f_k^*$ as, $$ f_k^*(\bm{x}) = f_k(P_k \bm{x}) $$ and $g^*$ as, $$ g^*(\bm{x}) = g(P\bm{x}) $$ It is easy to check that $f_k^*, \; k = 1,\ldots,K $ and $g^*$ are densities, as $ P_k $'s and $P$ are orthonormal matrices. Now, we can use $f^*_k$'s as our true speaker models, $g^*$ as the model obtained from the test utterance and obtain the intended speaker following the minimum distance based approach described previously. In particular for LD, we get the new modified equation from~(\ref{eq:loglik}) as,
\begin{equation}
\hat{k}^* = \argmax_k \sum\limits_{i = 1}^{M}\log(f^*_k(\bm{x}_i)) = \argmax_k \sum\limits_{i = 1}^{M}\log(f_k(P_k \bm{x}_i))
\end{equation}
which is the same as the PCT-based approach proposed in our previous work~\cite{pal2014}.
\setlength{\unitlength}{1in}

\begin{figure}[htb!]
\begin{center}
\framebox(6.2,1.6){
\begin{tikzpicture}[node distance=2cm]

\node (step1) [item_elp, align=center] {Training \\ Utterance};
\node (step2) [item_elp, align=center, xshift=3.6cm] {MFCC \\ Vectors};
\node (step2_0) [item_elp, align=center, xshift=7.2cm] {Computation \\ of PCT};
\node (step2_1) [item_elp, align=center, xshift=11.5cm] {PC-Transformed \\ MFCC Vectors};
\node (step3) [item_elp, align=center, yshift=-2cm, xshift=7cm] {Estimation of \\ Speaker GMM($f_k$)};
\node (step4) [item_rec, align=center, xshift=12.1cm,yshift=-2cm] {PCT, GMM database \\ for speaker};
\draw [arrow_t] (step1) -- (step2);
\draw [arrow_t] (step2) -- (step3);
\draw [arrow_t] (step2) -- (step2_0);
\draw [arrow_t] (step2_0) -- (step2_1);
\draw [arrow_t] (step3) -- (step4);
\draw [arrow_t] (step2_0) -- (step4);
\draw [arrow_t] (step2_1) -- (step3);

\end{tikzpicture}
}

\small(a) Training Module
\vspace{3mm}

\hbox

\framebox(6.2,1.8){
\begin{tikzpicture}[node distance=2cm]

\node (step2_1) [item_rec, align=center, xshift=12.1cm] {PCT, GMM database \\ for Speaker};
\node (step1) [item_elp, align=center, yshift=-1cm] {Test \\ Utterance};
\node (step2_0) [item_elp, align=center, xshift=3.6cm, yshift=-1cm] {MFCC \\ Vectors};
\node (step3) [item_elp, align=center, xshift=7.8cm, yshift=-1cm] {PCT-transformed \\ MFCC Vectors};
\node (step4) [item_elp, align=center, yshift=-3cm, xshift=7cm] {Estimation of \\ Test Utterance GMM};
\node (step5) [item_rec, align=center, xshift=12.4cm,yshift=-3cm] {Divergence \\ Measure};
\draw [arrow_t] (step1) -- (step2_0);
\draw [arrow_t] (step2_0) -- (step3);
\draw [arrow_t] (step2_1) -- node[anchor=south] {PCT} (step3);
\draw [arrow_t] (step3) -- (step4);
\draw [arrow_t] (step2_1) -- node[anchor=west] {GMM} (step5);
\draw [arrow_t] (step4) -- node[anchor=south] {GMM} (step5);

\end{tikzpicture}
}

\small(b) Test Module
\vspace{3mm}

\hbox

\framebox(6.2,2.4){
\begin{tikzpicture}[node distance=2cm]

\node (step1_0) [item_rec, align=center] {Classifier no. 1};
\node (step1_1) [item_rec, align=center, yshift=-1.2cm] {Classifier no. 2};
\node (step1_2) [item_rec, align=center, yshift=-2.4cm] {Classifier no. 3};
\node (step1_3) [item_rec, align=center, yshift=-3.6cm] {Classifier no. 4};
\node (step2_0) [item_rec, align=center, xshift = 4cm, yshift = 0.5cm] {Divergence from \\ Speaker Model no. 1};
\node (step2_1) [item_rec, align=center, xshift = 4cm, yshift = -1cm] {Divergence from \\ Speaker Model no. 2};
\node (step2_2) [item_rec1, align=center, xshift = 4cm, yshift = -2.5cm] {:};
\node (step2_3) [item_rec, align=center, xshift = 4cm, yshift = -4cm] {Divergence from \\ Speaker Model no. N};
\node (step3) [item_rec, align=center, xshift=8cm, yshift = -1.75cm] {Minimizer};
\node (step4) [item_rec, align=center, xshift=12cm, yshift = -1.75cm] {Classification};
\draw [arrow_t] (step1_0.east) -- (step2_1.west);
\draw [arrow_t] (step1_0.east) -- (step2_2.west);
\draw [arrow_t] (step1_0.east) -- (step2_3.west);
\draw [arrow_t] (step1_1.east) -- (step2_0.west);
\draw [arrow_t] (step1_1.east) -- (step2_1.west);
\draw [arrow_t] (step1_1.east) -- (step2_2.west);
\draw [arrow_t] (step1_1.east) -- (step2_3.west);
\draw [arrow_t] (step1_2.east) -- (step2_0.west);
\draw [arrow_t] (step1_2.east) -- (step2_1.west);
\draw [arrow_t] (step1_2.east) -- (step2_2.west);
\draw [arrow_t] (step1_2.east) -- (step2_3.west);
\draw [arrow_t] (step1_3.east) -- (step2_0.west);
\draw [arrow_t] (step1_3.east) -- (step2_1.west);
\draw [arrow_t] (step1_3.east) -- (step2_2.west);
\draw [arrow_t] (step1_3.east) -- (step2_3.west);
\draw [arrow_t] (step2_0.east) -- (step3.west);
\draw [arrow_t] (step2_1.east) -- (step3.west);
\draw [arrow_t] (step2_2.east) -- (step3.west);
\draw [arrow_t] (step2_3.east) -- (step3.west);
\draw [arrow_t] (step3.east) -- (step4.west);

\end{tikzpicture}
}

\small(c) Classifier Combination (using 4 classifiers)
 \vspace{5mm}

\caption{Flow charts for the three components of the proposed speaker identification method}
\end{center}
\end{figure}

Flow charts of the different components (training, testing and classifier combination) of the proposed approach are given in Figure 2.

\section{Implementation and Results}

The proposed approach was validated on two speech corpora, whose details are given in the following section.

\subsection{ISIS and NISIS: New Speech Corpora}
ISIS (an acronym for Indian Statistical Institute Speech) and NISIS (Noisy ISIS)~\cite{pal2012} are  speech corpora, which respectively contain simultaneously-recorded microphone and telephone speech of  105 speakers, over multiple sessions, spontaneous as well as read, in two languages  (Bangla and English), recorded in a typical office environment with moderate background noise. They were created in the Indian Statistical Institute, Kolkata, as a part of a project funded by the Department of Information Technology, Ministry of Communications and Information Technology, Government of India, during 2004-07. The speakers had Bangla or another Indian language as their mother tongue, and so were non-native English speakers.
Particulars of both corpora are given below:

\begin{itemize}
	\item Number of speakers	:	105 (53 male + 52  female)
\item	Recording environment:	moderately quiet computer room
\item	Sessions per speaker:	4 (numbered I, II, III and IV)
\item	Interval between sessions:	1 week to about 2 months
\item	Types of utterances in Bangla and English per session:	

\begin{itemize}
	\item 10 isolated words (randomly drawn from a specific text corpus, and generally different for all speakers and sessions)
\item	answers to 8 questions (these answers included dates, phone numbers, alphabetic sequences, and a few words spoken spontaneously)
\item	12 sentences (first two sentences common to all speakers, the remaining randomly drawn from the text corpus, duration ranging from 3-10 seconds)
\end{itemize}
\end{itemize}
Thus, for each session, there are two sets of recordings per speaker, one each in Bangla and English, containing 21 files each. 

\subsection{The Benchmark Telephone Speech Corpus NTIMIT}

NTIMIT~\cite{fisher1993,jankowski1990}, like TIMIT~\cite{fisher1986,garofolo1993}  is an acoustic-phonetic speech corpus in English, belonging to the Linguistic Data Consortium (LDC) of the University of Pennsylvania. TIMIT consists of clean microphone recordings of 10 different read sentences (2 \textit{sa}, 3 \textit{si} and 5 \textit{sx} sentences, some of which have rich phonetic variability), uttered by 630 speakers (438 males and 192 females) from eight major dialect regions of the USA.   It is characterized by 8-\textit{kHz} bandwidth and lack of intersession variability, acoustic noise, and microphone variability or distortion. These features make TIMIT a benchmark of choice for researchers in several areas of speech processing. 

NTIMIT, on the other hand, is the speech from the TIMIT database played through a carbon-button telephone handset and recorded over local and long-distance telephone loops. This provides speech identical to TIMIT, except that it is degraded through carbon-button transduction and actual telephone line conditions. Performance differences between identical experiments on TIMIT and NTIMIT are therefore, expected to arise primarily from the degrading effects of telephone transmission. Since the ordinary MFCC-GMM model achieves near perfect accuracy on TIMIT, further improvement seems to be unlikely. Therefore we have experimented with the NTIMIT database exclusively.

\subsection{Features Used}
\label{sec:feature}
The features used in this work are the widely-used Mel-frequency cepstral coefficients (MFCCs)~\cite{davis1980}, which  are coefficients that collectively make up a Mel Frequency Cepstrum (MFC). The latter is a representation of the short-time power spectrum of a sound signal, based on a linear cosine transform of a log-energy spectrum on a nonlinear mel scale of frequency. It exploits auditory principles, as well as the decorrelating property of the cepstrum, and is amenable to compensation for convolution distortion. As such, it has turned out to be one of the most effective feature representations in speech-related recognition tasks~\cite{quatieri2008}. A given speech signal is partitioned into overlapping segments or frames, and MFCCs are computed for each such frame.  Based on a bank of $K$ filters, a set of $M$ MFCCs is computed from each frame~\cite{pal2014}.

In addition, the  delta  Mel-frequency cepstral coefficients~\cite{quatieri2008}, which are nothing but the first-order frame-to-frame differences of the MFCCs, have also been used.
\subsection{Results}

The evaluation of the proposed method has been performed with the help of 10 recordings per speaker in both corpora, with the help of two different data sets:
\begin{itemize}
	\item  Dataset 6:4: consisting of the first 6 utterances for training and remaining 4 for testing
	\item Dataset 8:2: consisting of thefirst 8 utterances for training and remaining 2 for testing
\end{itemize}
In addition, evaluation has been done on two different sets of features: 
\begin{itemize}
	\item FS-I: 20 MFCCs and 20 delta MFCCs
	\item FS-II: 39 MFCCs
\end{itemize}
To implement the ensemble classification principle, on a number of competing MFCC-GMM classifiers were generated by varying certain tuning parameters of the generic MFCC-GMM classifier; the values of the parameters tuned (\textit{window size}, \textit{minimum frequency} and \textit{maximum frequency})  are mentioned in the tables. 
The accuracy of the aggregated GMM-MFCC classifier is obtained by combining the likelihood scores of the individual classifier components.  

The best performance  observed on NTIMIT in our earlier work~|cite{pal2014} has been summarized in Table~\ref{tab:results_prev} without the PCT (WOPCT) as well as with PCT (WPCT). These will be used as the baseline for assessing the efficacy of the proposed approach based on the Minimum Rescaled Modified  Distance Estimators (MRMDEs), employing all three divergence measures described in Section~\ref{sec:proposed}.
\subsection{Results with NTIMIT}
\label{sec:resntimit} 
Table~\ref{tab:ntimit} gives the identification accuracy on NTIMIT with the proposed approach, using all three divergence measures described in . From the latter it is evident that significant improvement has been achieved with MRMDEs based on all three divergence measures. Moreover, in each case, FS-I, which contains 20 MFCCs and 20 delta MFCCs, gives uniformly better performance than FS-II, consisting of 39 MFCCs only. Overall, the best performance of 56.19\% with the 6:4 dataset  and 67.86\% with the 8:2 dataset has been obtained with the LD divergence, using FS-I. These represent an improvement of over 10\% over the baseline performance.

\begin{table}[h]
\caption{Performance of the Baseline MFCC-GMM Speaker Identification system}
\label{tab:results_prev}
\begin{center}
\setlength{\extrarowheight}{1pt}
\begin{tabular}{|c|c|d{2cm}|d{2cm}|d{2cm}|d{2cm}|d{2cm}|}
\hline
\multirow{2}{*}{Corpus} &\multirow{2}{*}{Data set} & \multicolumn{2}{c|}{Individual} & \multicolumn{2}{c|}{Aggregate} \\
\cline{3-6}
&& WOPCT & WPCT & WOPCT & WPCT \\
\hline
\multirow{2}{*}{NTIMIT} &6:4 & 34.96 & 42.26 & 40.36 & 45.99 \\
\cline{2-6}
&8:2 & 42.41 & 52.30 & 49.05  & 55.63 \\ \hline
\multirow{2}{*}{\shortstack{NISIS\\(ES-I)}}&6:4 & 68.50 & 85.50 & 71.50 & 86.50 \\ \cline{2-6}
& 8:2 &76.00&89.00&77.00&91.50 \\
\hline
\end{tabular}
\end{center}
\end{table}
\begin{sidewaystable}[htb!]
\scriptsize
\begin{center}
{
\caption{Identification accuracy on NTIMIT under the proposed approach} 
\label{tab:ntimit}
\setlength{\extrarowheight}{1pt}
\begin{tabular}{|d{1cm}|d{1cm}|d{1cm}|d{1cm}|d{1cm}|d{1cm}|d{1cm}|d{1cm}|d{1cm}|d{1cm}|d{1cm}|d{1cm}|d{1cm}|d{1cm}|d{1cm}|}
\hline
\multirow{3}{*}{Dataset} & \multirow{3}{*}{\shortstack{Experi-\\ment}} & \multirow{3}{*}{\shortstack{Window\\Size\\(ms)}} & \multicolumn{4}{c|}{Based on $C_{MLD}(\delta)$} & \multicolumn{4}{c|}{Based on $C_{MHD}(\delta)$} & \multicolumn{4}{c|}{Based on $C_{MPCS}(\delta)$} \\
\cline{4-15}
& & & \multicolumn{2}{c|}{WOPCT} & \multicolumn{2}{c|}{WPCT} & \multicolumn{2}{c|}{WOPCT} & \multicolumn{2}{c|}{WPCT} & \multicolumn{2}{c|}{WOPCT} & \multicolumn{2}{c|}{WPCT} \\
\cline{4-15}
& & & FS-I & FS-II & FS-I & FS-II & FS-I & FS-II & FS-I & FS-II & FS-I & FS-II & FS-I & FS-II \\
\hline
\multirow{3}{*}{6:4} & 1 & 0.020 & \textbf{43.293} & \textbf{40.952} & 46.547 & 45.595 & \textbf{41.507} & \textbf{38.095} & \textbf{45.357} & \textbf{43.373} & 39.246 & \textbf{36.269} & 43.452 & 41.031 \\
\cline{2-15}
& 2 & 0.030 & 42.936 & 39.127 & \textbf{47.142} & \textbf{45.714} & 41.269 & 36.389 & 45.158 & 43.214 & \textbf{39.603} & 35.317 & \textbf{43.650} & \textbf{42.222} \\
\cline{2-15}
& \multicolumn{2}{c|}{Combined} & \multicolumn{2}{c|}{52.540} & \multicolumn{2}{c|}{56.190} & \multicolumn{2}{c|}{49.563} & \multicolumn{2}{c|}{53.730} & \multicolumn{2}{c|}{51.667} & \multicolumn{2}{c|}{53.889} \\
\hline
\multirow{3}{*}{8:2} & 1 & 0.020 & \textbf{56.031} & \textbf{52.381} & 59.523 & \textbf{57.539} & 53.571 & 49.603 & 57.539 & 54.761 & 51.587 & 46.587 & 55.159 & 50.555 \\
\cline{2-15}
& 2 & 0.030 & 56.270 & 49.365 & \textbf{60.079} & \textbf{57.539} & 54.444 & 46.666 & 57.301 & 55.317 & 52.142 & 44.365 & 56.349 & 53.253 \\
\cline{2-15}
& \multicolumn{2}{c|}{Combined} & \multicolumn{2}{c|}{64.524} & \multicolumn{2}{c|}{67.857} & \multicolumn{2}{c|}{61.429} & \multicolumn{2}{c|}{64.206} & \multicolumn{2}{c|}{63.571} & \multicolumn{2}{c|}{66.111} \\
\hline
\end{tabular}
\bigskip \bigskip \bigskip \bigskip

\caption{Identfication accuracy on  NISIS (ES-I) under the proposed approach}
\label{tab:nisis}
\setlength{\extrarowheight}{1pt}
\begin{tabular}{|d{1cm}|d{1cm}|d{1cm}|d{1cm}|d{1cm}|d{1cm}|d{1cm}|d{1cm}|d{1cm}|d{1cm}|d{1cm}|d{1cm}|d{1cm}|d{1cm}|d{1cm}|d{1cm}|d{1cm}|}
\hline
\multirow{3}{*}{Dataset} & \multirow{3}{*}{\shortstack{Experi-\\ment}} & \multirow{3}{*}{\shortstack{Min\\Freq\\(Hz)}} & \multirow{3}{*}{\shortstack{Max\\Freq\\(Hz)}} & \multirow{3}{*}{\shortstack{Window\\Size\\(ms)}} & \multicolumn{4}{c|}{Based on $C_{MLD}(\delta)$} & \multicolumn{4}{c|}{Based on $C_{MHD}(\delta)$} & \multicolumn{4}{c|}{Based on $C_{MPCS}(\delta)$} \\
\cline{6-17}
& & & & & \multicolumn{2}{c|}{FS-I} & \multicolumn{2}{c|}{FS-II} & \multicolumn{2}{c|}{FS-I} & \multicolumn{2}{c|}{FS-II} & \multicolumn{2}{c|}{FS-I} & \multicolumn{2}{c|}{FS-II} \\
\cline{6-17}
& & & & & WOPCT & WPCT & WOPCT & WPCT & WOPCT & WPCT & WOPCT & WPCT & WOPCT & WPCT & WOPCT & WPCT \\
\hline
\multirow{4}{*}{6:4} & 1 & 200 & 4000 & 0.020 & 83.25 & 88 & 81.75 & 85.5 & 82.25 & 86.5 & 78.75 & 83.5 & 79 & 85 & 78.5 & 81.75 \\
\cline{2-17}
& 2 & 200 & 4000 & 0.030 & 83.75 & 86.75 & 79.5 & 85.5 & 82.25 & 84.25 & 76.25 & 83.25 & 80 & 84.25 & 75.5 & 82.25 \\
\cline{2-17}
& 3 & 0 & 5500 & 0.020 & 86.5 & \textbf{89.75} & 82.75 & 85.75 & 84.5 & \textbf{89.5} & 81.75 & 84 & \textbf{83} & \textbf{88.75} & \textbf{82.5} & 84.5 \\
\cline{2-17}
& 4 & 0 & 5500 & 0.030 & \textbf{87.75} & 89 & \textbf{83} & \textbf{87.75} & \textbf{86} & 87.75 & \textbf{82} & \textbf{86} & 82.75 & 87 & 81 & \textbf{84.75} \\
\cline{2-17}
& \multicolumn{4}{c|}{1-4 Combined} & 87.5 & 92 & 84.5 & 88.5 & 86 & 89.5 & 83 & 86.5 & 85.5 & 88.75 & 83.75 & 86.5 \\
\hline
\multirow{4}{*}{8:2} & 1 & 200 & 4000 & 0.020 & 88.5 & 93 & 85 & \textbf{91.5} & 86.5 & 90.5 & 85 & 88.85 & 86 & 91 & 83.5 & 86.5 \\
\cline{2-17}
& 2 & 200 & 4000 & 0.030 & 89.5 & 93 & 85.5 & \textbf{91.5} & 87 & 90.5 & 82.5 & 89 & \textbf{88} & 91 & 82 & 88 \\
\cline{2-17}
& 3 & 0 & 5500 & 0.020 & 90 & \textbf{94.5} & 89 & \textbf{91.5} & 86 & \textbf{92} & 87.5 & 90 & 87.5 & \textbf{91.5} & \textbf{86.5} & 89 \\
\cline{2-17}
& 4 & 0 & 5500 & 0.030 & \textbf{90.5} & 92.5 & \textbf{89.5} & 93 & \textbf{89} & 90.5 & \textbf{88} & \textbf{90.5} & \textbf{88} & 90.5 & 86 & \textbf{91} \\
\cline{2-17}
& \multicolumn{4}{c|}{1-4 Combined} & 90 & 94.5 & 92.5 & 93.5 & 88 & 92.5 & 88 & 92.5 & 89.5 & 91.5 & 90 & 91.5 \\
\hline
\end{tabular}
}
\end{center}
\end{sidewaystable}

\subsection{Results with NISIS}

The best performance  observed on NISIS using English recordings from Session I only (referred to as ES-I) in our earlier work (Bose \textit{et al.}, 2014) has been summarized in Table~\ref{tab:results_prev} without the PCT (WOPCT) as well as with PCT (WPCT), while Table~\ref{tab:nisis} gives the identification accuracy on it with the proposed approach, using all three divergence measures described in Section~\ref{sec:proposed}. As in the case of NTIMIT, it is seen  that significant improvement has been achieved with MRMDEs in each divergence measure. Moreover, as observed earlier with NTIMIT, FS-I  gives uniformly better performance than FS-II, in each instance. Again, as before~\ref{sec:resntimit} he best overall performance of 92\% with the 6:4 dataset  and 94.5\% with the 8:2 dataset has been obtained with the LD divergence. These represent an improvement of about 6\% over the baseline performance.

It is worth noting that the improvement on NISIS is not as dramatic as that with NTIMIT. The explanation is that, the baseline performance  with NISIS being quite high to begin with, there is not too much scope for improving that further. This may possibly be another positive feature of the proposed approach, namely, its ability to provide a relatively stronger boost to weaker baseline methods.

\section{Conclusions}
In the usual approach of Speaker identification, the probability distribution of the MFCC features for each speaker is modeled using Gaussian Mixture Models. For a test utterance, its MFCC feature vectors are matched with the speaker models using the likelihood scores derived from each model.  The test utterance is assigned to the model with highest likelihood score.

In this work, a novel solution to the speaker identification problem is proposed through minimization of statistical divergences between the probability distribution ($g$) of feature vectors derived from the test utterance and the probability distributions of the feature vectors corresponding to the speaker classes. This approach is made more robust to the presence of outliers, through the use of suitably modified versions of the standard divergence measures.  Three such measures were considered -- the likelihood disparity, the Hellinger distance and the Pearson chi-square distance. 

It turns out that the proposed approach with the likelihood disparity, when the empirical distribution function is used to estimate $g$, becomes equivalent to maximum likelihood classification with Gaussian Mixture Models (GMMs) for speaker classes, the usual approach discussed above. The usual approach was used for example, by Reynolds (1995) yielding excellent results. Significant improvement in classification accuracy is observed under the current approach on the benchmark speech corpus NTIMIT and a new bilingual speech corpus NISIS, with MFCC features, both in isolation and in combination with delta MFCC features. 
Further, the ubiquitous principal component transformation, by itself and in conjunction with the principle of classifier combination, improved the performance even further.  

\section{Acknowledgement}
The authors gratefully acknowledge the contribution of Ms Disha Chakrabarti and Ms Enakshi Saha to this work.
\bibliography{references}

\begin{thebibliography}{10}

\bibitem{altincay2003}
H.~Altin\c{c}ay and M.~Demirekler.
\newblock Speaker identification by combining multiple classifiers using
  {D}empster-{S}hafer theory of evidence.
\newblock {\em Speech Communication}, 41:531--547, 2003.

\bibitem{basu2011}
Ayanendranath Basu, Hiroyuki Shioya, and Chanseok Park.
\newblock {\em Statistical Inference: The Minimum Distance Approach}.
\newblock CRC Press, Boca Raton, FL, 2011.

\bibitem{besacier2000}
L.~Besacier and J.-F. Bonastre.
\newblock Subband architecture for automatic speker recognition.
\newblock {\em Signal Processing}, 80:1245--1259, 2000.

\bibitem{breiman1996}
L.~Breiman.
\newblock Bagging predictors.
\newblock {\em Machine Learning}, 24:123--140, 1996.

\bibitem{breiman2001}
L.~Breiman.
\newblock Random forests.
\newblock {\em Machine Learning}, 45(1):5--32, 2001.

\bibitem{campbell1997}
J.~P. {Campbell (Jr.)}.
\newblock Speaker recognition: a tutorial.
\newblock {\em Proceedings of the {IEEE}}, 85:1437--1462, 1997.

\bibitem{chien2004}
J-T Chien and C-W Ting.
\newblock Speaker identificaton using probabilistic {PCA} model selection.
\newblock In {\em In {INTERSPEECH}-2004--{ICSLP},8th International Conference
  on Spoken Language Processing}, pages 1785--1788, Jeju Island, Korea, October
  4-8 2004.

\bibitem{davis1980}
S.~B. Davis and P.~Mermelstein.
\newblock Comparison of parametric representations for monosyllabic word
  recognition in continuously spoken sentences.
\newblock {\em {IEEE} Transactions on Acoustics, Speech, and Signal
  Processing}, 28(4):357--–366, 1980.

\bibitem{fisher1986}
W.M. Fisher, G.R. Doddington, and K.M. Goudie-Marshall.
\newblock The {DARPA} speech recognition research database:specifications and
  status.
\newblock {\em {DARPA} Workshop Speech Recognition}, pages 93--99, 1986.

\bibitem{fisher1993}
W.M. Fisher, G.R. Doddington, K.M. Goudie-Marshall, C.~Jankowski,
  A.~Kalyanswamy, S.~Basson, and J.~Spitz.
\newblock {NTIMIT}.
\newblock Linguistic Data Consortium, Philadelphia, 1993.

\bibitem{furui1997}
S.~Furui.
\newblock Recent advances in speaker recognition.
\newblock {\em Pattern Recognition Letters}, 18:859--872, 1997.

\bibitem{garofolo1993}
J.S. Garofolo, L.F. Lamel, W.M. Fisher, J.G. Fiscus, D.S. Pallett, N.L.
  Dahlgren, and V.~Zue.
\newblock {TIMIT} acoustic-phonetic continuous speech corpus.
\newblock Linguistic Data Consortium, Philadelphia, 1993.

\bibitem{hanilci2009}
C.~Hanil\c{c}i and F.~Erta\c{s}.
\newblock Principal component based classification for text-independent speaker
  identification.
\newblock In {\em Fifth International Conference on Soft Computing, Computing
  with Words and Perceptions in System Analysis,Decision and Control}, pages
  1--4, 2009.

\bibitem{jankowski1990}
C.~Jankowski, A.~Kalyanswamy, S.~Basson, and J.~Spitz.
\newblock {NTIMIT}: a phonetically balanced, continuous speech, telephone
  bandwidth speech database.
\newblock In {\em International Conference on Acoustics, Speech, and Signal
  Processing ({ICASSP}-90)}, 1990.

\bibitem{kinnunen2010}
T.~Kinnunen and H.~Li.
\newblock An overview of text-independent speaker recognition: from features to
  supervectors.
\newblock {\em Speech Communication}, 52:12--40, 2010.

\bibitem{vijendra2013}
D.~Vijendra Kumar, K.~Jyoti, V.~Sailaja, and N.M.~Ramalingeswara Rao.
\newblock Text independent speaker identification with principal component
  analysis.
\newblock {\em International Journal of Innovative Research in Science,
  Engineering and Technology}, 2:4433--4440, 2013.

\bibitem{lindsay1994}
B.~G. Lindsay.
\newblock Efficiency versus robustness: The case for minimum {H}ellinger
  distance and related methods.
\newblock {\em Ann. Statist.}, 22:1081--1114, 1994.

\bibitem{pal2014}
A.~Pal, S.~Bose, G.~K. Basak, and A.~Mukhopadhyay.
\newblock Speaker identiﬁcation by aggregating {G}aussian mixture models
  ({GMM}s) based on uncorrelated {MFCC}-derived features.
\newblock {\em International Journal of Pattern Recognition and Artiﬁcial
  Intelligence}, 28(4), 2014.

\bibitem{pal2012}
A.~Pal, S.~Bose, M.~Mitra, and S.~Roy.
\newblock {ISIS} and {NISIS}: New bilingual dual-channel speech corpora for
  robust speaker recognition.
\newblock In {\em the 2012 International Conference on Image Processing,
  Computer Vision and Pattern Recognition (IPCV 2012)}, pages 936--939, Las
  Vegas, USA, 2012.

\bibitem{quatieri2008}
T.F. Quatieri.
\newblock {\em Discrete-Time Speech Signal Processing: Principles and
  Practice}.
\newblock Pearson Education, Inc., 2008.

\bibitem{rao2001}
C.R. Rao.
\newblock {\em Linear Statistical Inference and Its Applications}.
\newblock John Wiley \& Sons, New York, 2nd (reprint) edition, 2001.

\bibitem{reynolds1995}
D.A. Reynolds.
\newblock Large population speaker identification using clean and telephone
  speech.
\newblock {\em {IEEE} Signal Processing Letters}, 2:46--48, 1995.

\bibitem{seo2001}
C.~Seo, K.Y. Lee, and J.~Lee.
\newblock {GMM} based on local {PCA} for speaker identification.
\newblock {\em Electronics Letters}, 37:1486--1488, 2001.

\bibitem{suri2013}
K.~{Suri Babu}, Y.~Anitha, and K.K.V.S. Anjana.
\newblock Dimensionality reduction in feature vector using principal component
  analysis ({PCA}) for effective speaker recognition.
\newblock {\em International Journal of Applied Information Systems}, 5:15--17,
  2013.

\bibitem{trabelsi2013}
I.~Trabelsi and D.~Ben Ayed.
\newblock A multi level data fusion approach for speaker identification on
  telephone speech.
\newblock {\em International Journal of Signal Processing, Image Processing and
  Pattern Recognition}, 6:33--41, 2013.

\bibitem{zhang2003}
W.~Zhang, Y.~Yang, and Z.~Wu.
\newblock Exploiting {PCA} classifiers to speaker recognition.
\newblock In {\em Proceedings of Int. Joint Conference on Neural Networks},
  pages 820--823, 2003.

\end{thebibliography}

\end{document}